\documentclass[runningheads]{llncs}
\usepackage[T1]{fontenc}
\usepackage{lmodern}
\usepackage{booktabs}
\usepackage{graphicx}
\usepackage[protrusion=true,expansion=false]{microtype}
\usepackage{amsmath,amssymb}
\usepackage{multirow}
\usepackage{siunitx}
\usepackage{cite}
\usepackage{xcolor}
\usepackage{needspace}
\usepackage{placeins}
\usepackage{float}
\usepackage[hidelinks,breaklinks=true]{hyperref}
\usepackage[all]{hypcap}
\usepackage[nameinlink,capitalise,noabbrev]{cleveref}
\begin{document}
\title{Candidate-Expanding Routing with Permutation-Stabilized Experts for Mixed-Format Medical VQA}
\titlerunning{Candidate-Expanding Routing for Medical VQA}
\author{Hai-Dang Nguyen\inst{1,2} \and
Huy-Hieu Pham\inst{1,2,3}\thanks{Corresponding author.}}
\authorrunning{H.-D. Nguyen and H.-H. Pham}
\institute{College of Engineering and Computer Science, VinUniversity, Hanoi, Vietnam \and
VinUni-Illinois Smart Health Center, VinUniversity, Hanoi, Vietnam \and
Center for Innovations in Health Sciences, VinUniversity, Hanoi, Vietnam\\
\email{dang.nh3@vinuni.edu.vn, hieu.ph@vinuni.edu.vn}}
\maketitle

\begin{abstract}
Mixed-format medical visual question answering (VQA) requires stable option selection and machine-readable free-text output. The two formats fail differently: multiple-choice predictions can change with option symbols or positions, while clinically plausible open answers can fail automated evaluation when serialization is malformed. We address both challenges with an answer-text memory, a permutation-stabilized vision--language expert, and a sparse candidate-expanding router. The cyclic schedule follows prior work; our contribution is to make expert top-2 a routable candidate alongside memory and expert top-1. On a 1,403-case retrospective internal analysis, this expansion improves a matched binary router from 88.95\% to \textbf{91.73\%} (+2.78 percentage points; 95\% CI 1.57--3.99), with 56 rescued errors and 17 regressions. Oracle coverage rises from 90.31\% to 96.15\%, and the final submitted configuration reaches \textbf{92.23\%} on the same retrospective split. For open questions, strict generation and deterministic guards produce 475/475 schema-valid participant-facing outputs without repair, retry, or hard-gate failure. Visual ablations reveal substantial textual dependence. Candidate expansion supplies the principal controlled routing gain; open-path evidence establishes output-contract validity rather than clinical correctness in medical use or deployment.
\end{abstract}

\section{Introduction}
Medical visual question answering (VQA) connects heterogeneous clinical images with questions ranging from recognition to multi-step reasoning~\cite{lin2023medicalvqa,li2023llavamed}. Recent public MLLMs include Qwen2.5-VL, MedGemma, and Lingshu~\cite{bai2025qwen25vl,sellergren2025medgemma,lasa2025lingshu}. Med-CMR spans 11 organ systems, 12 modalities, and seven visual or clinical-reasoning tasks~\cite{gong2026medcmr}. Its mixed format creates two interface problems: MCQs require stable selection from case-defined options, while open questions require useful content in a machine-readable form. An MCQ can fail through unstable option identifiers, whereas a clinically plausible open answer can become unscorable through malformed fields. A single submission must control both semantic selection and the observable output contract.

MCQ selection is sensitive to option order and symbols~\cite{pezeshkpour2024order,yang2025symbol}. Prior cyclic evaluation exposes each semantic option to every identifier before back-mapping and aggregation~\cite{zheng2024robust}; permutation sensitivity also affects VLMs~\cite{zong2024permutations}, motivating recent LVLM bias correction~\cite{atabuzzaman2025selection}. Calibration estimates label priors~\cite{zhao2021calibrate}, and self-consistency aggregates repeated reasoning paths~\cite{wang2023selfconsistency}. Rather than collapse repeated scores to one answer, we retain ranked semantic alternatives so that a correct top-2 option remains available to the selector.

Retrieval-augmented and medical memory models usually condition one generator on retrieved evidence~\cite{lewis2020rag,yuan2023ramm}; learning-to-defer and selective-prediction methods instead choose which predictor should answer~\cite{mozannar2020defer,mao2025routing,geifman2017selective}. We keep memory and VLM as independent candidate generators and merge them only at selection. Routing memory with expert top-1 and top-2 expands reachable answers when the first two candidates are wrong. It also separates \emph{candidate coverage}, whether any candidate is correct, from \emph{router regret}, errors made after a correct candidate is available.

For open VQA, hallucination and rationale faithfulness concern clinical content and causal support~\cite{gu2024medvh,jacovi2020faithfulness}; valid JSON establishes neither. Parsing, deterministic guards, and one retry address malformed fields, hidden text, and repetition only as an output contract, without claiming clinical verification.

To address these challenges, we separate candidate formation from selection. The MCQ path constructs one memory candidate and two ranked VLM candidates; the open path places one structured generation behind an explicit serialization contract. The main contributions are:
\begin{itemize}
\item We formulate MCQ inference as heterogeneous candidate expansion over answer-text memory and expert top-1/top-2. The third candidate raises matched accuracy by 2.78 percentage points and oracle coverage by 5.84 points on retrospective internal analysis.
\item We adapt prior cyclic scheduling to case-defined medical-VQA labels through token-aware scoring, semantic back-mapping, and option-space aggregation while retaining ranked expert outputs for selection.
\item We separate coverage from routing error through matched ablations, rescue and regression counts, and router regret. Shortcut controls and the open-output audit bound the resulting claims.
\end{itemize}
The rest of the paper is organized as follows. \Cref{sec:method} presents the permutation-stabilized expert, answer-text memory, candidate-expanding router, and open-output contract. \Cref{sec:experiments} describes the experimental protocol, baselines, controlled ablations, mechanistic analyses, qualitative evidence, and robustness studies. \Cref{sec:discussion} discusses findings, limitations, and conclusions.

\clearpage
\section{Method}
\label{sec:method}
\subsection{System Overview}
\begin{figure}[!htbp]
\centering
\includegraphics[width=0.95\textwidth]{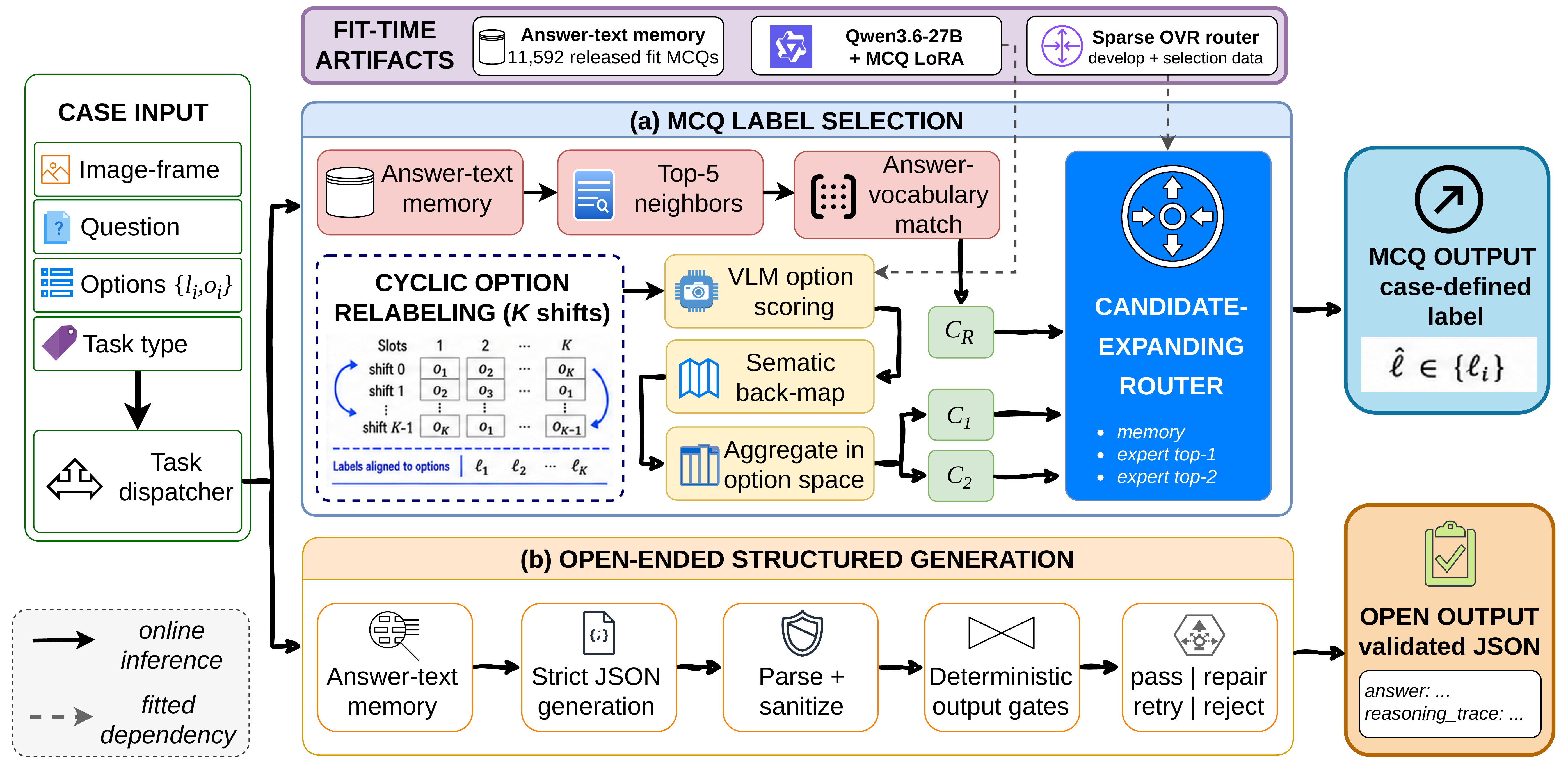}
\caption{Mixed-format input-to-output pipeline. A shared image and question are dispatched by task type. (a) For MCQs, answer-text memory independently yields $c_R$, while cyclic relabeling, VLM scoring, and semantic back-mapping yield $c_1$ and $c_2$; our router returns one valid case-defined label. (b) For open questions, retrieved context conditions strict JSON generation, followed by parsing and deterministic gates that return validated \texttt{answer} and \texttt{reasoning\_trace} fields or trigger repair, retry, or rejection. Dashed arrows are fitted dependencies; solid arrows are online inference.}
\label{fig:pipeline}
\end{figure}

The common input is a medical image $I$, question $q$, task type, and, for MCQ, case-defined label--text pairs $O=\{(\ell_i,o_i)\}_{i=1}^{K}$. A \emph{slot} joins a displayed symbol and its position. The MCQ path outputs one $\hat\ell\in\{\ell_i\}$; the open path outputs a validated JSON object containing \texttt{answer} and \texttt{reasoning\_trace}. \Cref{fig:pipeline} separates these paths and organizes each as input, method, and output. MCQ retrieval and VLM scoring share no prediction state and meet only at the router; the open branch uses its own retrieval and generation path.

\FloatBarrier

\subsection{Heterogeneous MCQ Candidate Generation}
\label{sec:retrieval}
\paragraph{Answer-text memory.}
The memory contains 11,592 released fit MCQs. Fit documents concatenate question, options, organizer visual description, metadata, and correct answer text; queries contain only question and options. Case-folded regex unigrams use smoothed IDF, L2 normalization, and cosine similarity ($\mathrm{min\_df}=1$). We rank 50 neighbors, retain five, match their answer text to current options, and similarity-vote for $c_R$. Historical labels and evaluated references are never transferred. Because fit descriptions contain a normalized answer substring in 422 cases, we call this an \emph{answer-aware answer-vocabulary memory} and audit the confound in \cref{sec:robustness}.

\paragraph{Permutation-stabilized expert.}
\label{sec:cyclic}
Following the cyclic selector evaluation of Zheng et al.~\cite{zheng2024robust}, we use a balanced $K$-shift schedule so that every semantic option occupies every case-defined symbol--position slot once. For $s\in\{0,\ldots,K-1\}$,
\begin{equation}
 \pi_s(i)=1+((i-s-1)\bmod K).
 \label{eq:perm}
\end{equation}
Only the slot assignment changes; image, question, and option content remain fixed. Option $o_i$ occupies $\ell_{\pi_s(i)}$; scores are mapped back before the submitted log-score average,
\begin{equation}
 \bar z_i=K^{-1}\sum_{s=0}^{K-1}z_s(\ell_{\pi_s(i)}),\qquad
 \hat i=\arg\max_i\bar z_i ,
 \label{eq:average}
\end{equation}
This ranks the geometric mean after within-shift normalization; its two highest scores define $c_1,c_2$. Under the additive slot model
$z_s(\ell_{\pi_s(i)})=u_i+b_{\pi_s(i)}+\epsilon_{is}$, balanced exposure contributes the same mean slot term to each option, but cannot remove content--slot interactions. Benchmark labels are single-token A--E; scoring tests six whitespace/punctuation variants. Parser/router tests also cover numeric, Roman, and arbitrary labels; ties follow case order.

\subsection{Candidate-Expanding Router}
\label{sec:router}
Candidates are memory $c_R$ and expert ranks $c_1,c_2$. A one-vs-rest L1 logistic model uses 58 inference features covering agreement, retrieval, confidence, input length, metadata, keywords, and shift consistency. References, correctness, IDs, and oracle switches are excluded. Targets choose the first correct candidate in priority $c_R,c_1,c_2$ and omit unreachable rows. Five-fold record-level out-of-fold predictions select class multipliers. The final \texttt{liblinear} fit ($C=1$, balanced classes, seed 26) uses development plus a permitted 100-case selection set. Inference applies factors $(1.0,1.0,0.7)$, masks duplicate/invalid candidates, and returns the exact case label.

\subsection{Open-Ended Output Contract}
Open retrieval searches 14,308 fit cases; up to three contexts enter one greedy JSON prompt (256-token cap; penalty 1.05; no repeated 8-grams). Both fields are generated once. Parsing accepts JSON or labeled fields, removes hidden text and Markdown, and caps answer/rationale at 35/80 words. Schema, length, repetition, symbol, meta-text, type, and consistency rules allow serialization-only repair and one retry; persistent failure aborts. Each object thus contains a nonempty answer and brief rationale. These checks verify output-contract properties, not clinical correctness or faithfulness, and are not hallucination detection.

\clearpage
\section{Experiments}
\label{sec:experiments}
\subsection{Experimental Setup}
\paragraph{Data and protocol.}
Med-CMR reports 20,653 VQA pairs across 11 organ systems, 12 modalities, and seven tasks~\cite{gong2026medcmr}. We stratify 17,722 labeled challenge cases~\cite{medreason2026guide} by question type, task, organ, and modality: fit has 14,308 cases (11,592 MCQ/2,716 open), while development and internal analysis each have 1,707 (1,403/304). Development supports selection. The later-inspected internal split is retrospective development evidence, not an independent evaluation. The unscored participant set has 2,057/475 cases and no final hidden result. MCQs use exact-label accuracy, 10,000 pHash-cluster bootstrap replicates~\cite{efron1993bootstrap}, and Holm-corrected exact McNemar tests~\cite{mcnemar1947note}. Patient/study IDs are unavailable. Med-CMR is CC BY-NC 4.0; no new patient data were collected, and use is research-only.

\paragraph{Models and baselines.}
Qwen3.6-27B revision \texttt{6a9e13b}~\cite{qwen2026qwen36} runs bfloat16 with a frozen vision stack. LoRA ($r=8$, $\alpha=16$, dropout 0.05) uses AdamW ($10^{-5}$), batch 1, accumulation 16/8, and seed 42 on one H100 80GB~\cite{hu2022lora}. A 300-step fit uses 11,592 MCQs plus 2,000 PMC-VQA cases with two relabelings~\cite{zhang2023pmcvqa}; 40-step continuation on a permitted 100-case selection set chooses step 30.

\subsection{Main Baselines}
\label{sec:baselines}
\Cref{tab:baselines} reports the reviewed same-split baselines. Candidate expansion raises binary routing from 88.95\% to 91.73\% (95\% CI $[1.57,3.99]$). The final retrospective result is 92.23\%; the development-tuned margin switch reaches 80.40\%.

\begin{table}[H]
\centering
\caption{Same-split baselines on 1,403 retrospective internal MCQs. The margin switch is selected on development and frozen; \emph{(Ours)} identifies the proposed and final systems.}
\label{tab:baselines}
\renewcommand{\arraystretch}{1.02}
\begin{tabular*}{\textwidth}{@{\extracolsep{\fill}}p{0.36\textwidth}S[table-format=2.2]p{0.36\textwidth}S[table-format=2.2]@{}}
\toprule
Method & {Acc. (\%)} & Method & {Acc. (\%)} \\
\midrule
\multicolumn{4}{@{}l}{\emph{Text / retrieval baselines}} \\
Global answer-frequency prior & 23.16 & Question-only retrieval & 35.71 \\
Answer-free lexical memory & 60.73 & Answer-aware lexical memory & 61.15 \\
\addlinespace[2pt]
\multicolumn{4}{@{}l}{\emph{VLM expert baselines}} \\
Base Qwen3.6 + context, direct & 58.95 & Label-shuffle LoRA, direct & 72.70 \\
Cyclic probability mean & 80.04 & Continued adapted expert & 81.25 \\
\addlinespace[2pt]
\multicolumn{4}{@{}l}{\emph{Routing systems}} \\
Dev-tuned margin switch & 80.40 & Memory + expert top-1 & 88.95 \\
\textbf{Top-2 router (Ours)} & \bfseries 91.73 & \textbf{Final system (Ours)} & \bfseries 92.23 \\
\bottomrule
\end{tabular*}

\end{table}

\subsection{Controlled Ablation}
\label{sec:ablation}
\Cref{tab:ablation} separates expert scoring from candidate expansion. Top-2 rescues 56 binary-router errors and causes 17 regressions: 39 net correct cases (+2.78 percentage points; 95\% CI $[1.57,3.99]$; Holm-adjusted $p=2.1\times10^{-5}$). Permutation diagnostics then change accuracy by $-0.14$ points (95\% CI $[-0.50,0.21]$).

\begin{table}[H]
\centering
\caption{Matched internal-MCQ changes. Each row states its comparison; CIs use pHash-cluster bootstrap sampling. \emph{(Ours)} marks candidate expansion. The base-to-adapted contrast confounds domain supervision with label-shuffle augmentation.}
\label{tab:ablation}
\renewcommand{\arraystretch}{1.00}
\begin{tabular*}{\textwidth}{@{\extracolsep{\fill}}p{0.43\textwidth}S[table-format=2.2]S[table-format=2.2]S[table-format=+2.2]r@{}}
\toprule
Controlled change & {From} & {To} & {$\Delta$ pp} & \multicolumn{1}{c}{95\% CI} \\
\midrule
\multicolumn{5}{@{}l}{\emph{A. Expert stabilization}} \\
Add cyclic stabilization to base & 58.95 & 63.36 & +4.42 & \makebox[7.4em][r]{[+2.36, +6.48]} \\
Add label-shuffle adaptation & 58.95 & 72.70 & +13.76 & \makebox[7.4em][r]{[+11.91, +15.67]} \\
Add cyclic plurality to adapted & 72.70 & 79.83 & +7.13 & \makebox[7.4em][r]{[+5.41, +8.92]} \\
Use probability-mean aggregation & 79.83 & 80.04 & +0.21 & \makebox[7.4em][r]{[-0.78, +1.21]} \\
Switch to submitted log mean & 80.04 & 79.90 & -0.14 & \makebox[7.4em][r]{[-0.79, +0.50]} \\
Continue adapted expert & 79.90 & 81.25 & +1.35 & \makebox[7.4em][r]{[+0.71, +2.06]} \\
\addlinespace[3pt]
\multicolumn{5}{@{}l}{\emph{B. Controlled router ablation}} \\
\textbf{Add expert top-2 (Ours)} & 88.95 & \bfseries 91.73 & \bfseries +2.78 & \textbf{\makebox[7.4em][r]{[+1.57, +3.99]}} \\
Add permutation diagnostics & 91.73 & 91.59 & -0.14 & \makebox[7.4em][r]{[-0.50, +0.21]} \\
\bottomrule
\end{tabular*}

\end{table}

\subsection{Mechanistic Analysis}
\label{sec:mechanism}
\begin{figure}[H]
\centering
\includegraphics[width=0.95\textwidth]{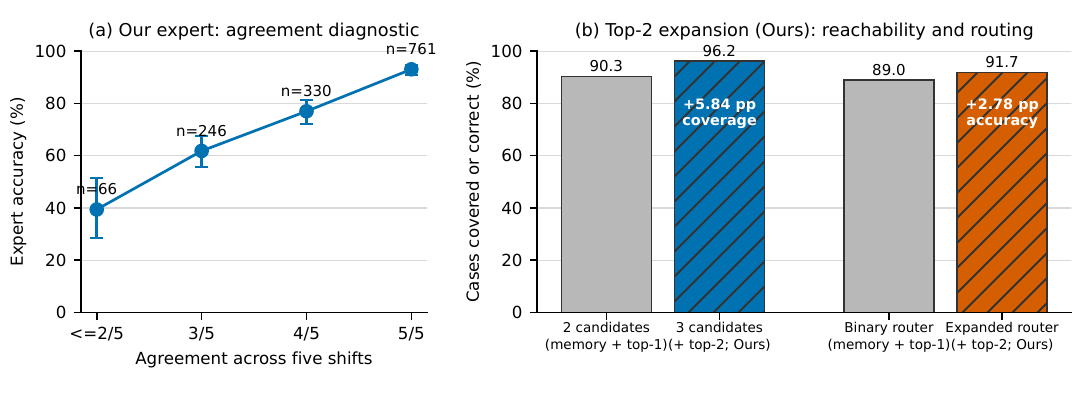}
\caption{Mechanistic evidence. (a) Expert accuracy versus five-shift agreement, with Wilson intervals and bin counts; agreement is diagnostic only. (b) Top-2 expands oracle coverage from 90.31\% to 96.15\% and matched routing from 88.95\% to 91.73\%; hatching marks expanded systems.}
\label{fig:quant}
\end{figure}

\Cref{fig:quant} distinguishes agreement-based diagnosis from candidate reachability. Adding $c_2$ raises oracle coverage from 90.31\% to 96.15\% and matched routing from 88.95\% to 91.73\%.

Cyclic scoring improves both checkpoints. Probability and log averaging are equivalent on identical logits (80.04\% vs. 79.90\%, $p=0.83$), while the adaptation contrast remains confounded by a missing same-data standard-LoRA checkpoint. Candidate expansion makes 82 answers newly reachable. The router selects $c_R/c_1/c_2$ for 861/427/115 cases, with conditional accuracies 96.86/93.68/52.17\%; $c_2$ cases are harder because earlier candidates conflict. The remaining 55 reachable errors yield 3.92 points of router regret. Agreement predicts correctness (AUROC 0.750), but margin is stronger (0.858), matching the neutral feature ablation in routing.

\subsection{Qualitative Evidence Across Formats}
\begin{figure}[H]
\centering
\includegraphics[width=0.96\textwidth]{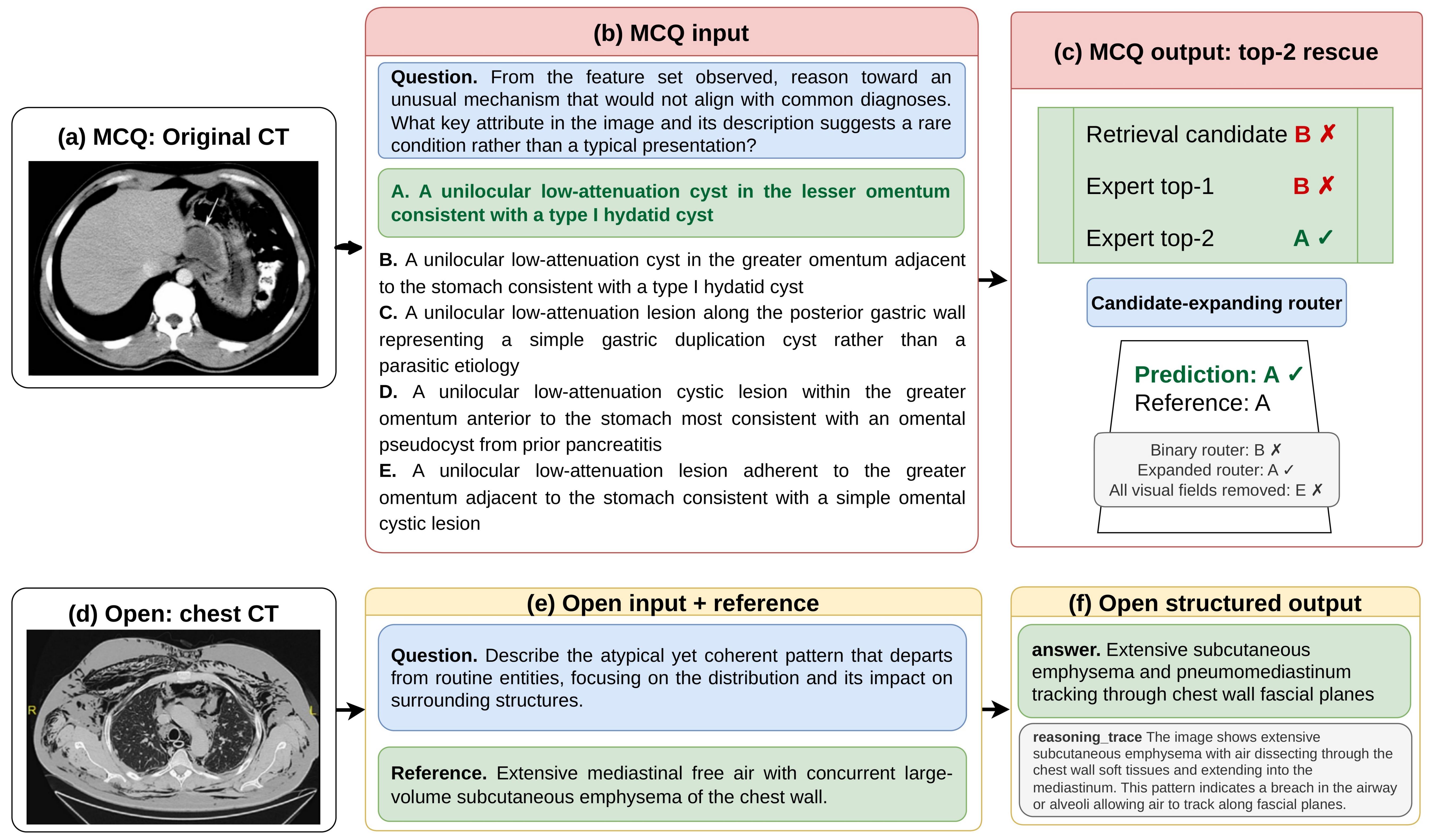}
\caption{Real Med-CMR input-to-output examples~\cite{gong2026medcmr,medreason2026guide}. (a--c) For abdominal CT, memory/top-1 predict B, top-2 exposes reference A, and our router selects A; visual removal predicts E. (d--f) For chest CT, the open output preserves both reference findings in valid JSON. Images are only resized; both are SHA-, pHash-, and template-disjoint from fit. These cases illustrate mechanism, not aggregate validity.}
\label{fig:rescue}
\end{figure}

\Cref{fig:rescue} shows two CT cases selected by fixed, reproducible criteria. In the MCQ, memory and top-1 fail; top-2, our router, and the reference agree, while visual removal produces another error. The open example maximizes a fixed answer/rationale overlap score among fit-disjoint radiology outputs and preserves both reference findings. Together they expose the two mechanisms without replacing aggregate or clinical validation.

\subsection{Robustness and Sensitivity}
\label{sec:robustness}
Options-only and question-plus-options retrieval score 61.30\% and 61.65\%, exposing answer-vocabulary regularity. Answer-free memory scores 60.73\% versus 61.15\% answer-aware; their disjoint scores are 60.35/61.04\%. We find 78 exact-image overlaps but no patient/study IDs. A pHash-grouped refit reaches 92.37\%, close to 92.23\%; this limits grouping sensitivity but is not independent validation. Each control in \cref{tab:visual} recomputes expert features and routing with the trained system fixed.

\begin{table}[H]
\centering
\caption{Visual sensitivity on retrospective internal MCQs. \emph{(Ours)} is matched input; other rows are independent controls, not a cumulative sequence.}
\label{tab:visual}
\renewcommand{\arraystretch}{1.03}
\begin{tabular*}{\textwidth}{@{\extracolsep{\fill}}p{0.30\textwidth}S[table-format=2.2]S[table-format=2.2]S[table-format=-2.2]p{0.17\textwidth}S[table-format=2.2]@{}}
\toprule
Input condition & {Expert} & {Router} & {$\Delta$ (pp)} & 95\% CI & {Flip} \\
\midrule
\textbf{Full input (Ours)} & \bfseries 81.25 & \bfseries 92.23 & 0.00 & \multicolumn{1}{c}{reference} & 0.00 \\
Raw image omitted & 76.27 & 88.95 & -3.28 & [-4.49, -2.14] & 6.91 \\
All visual fields omitted & 75.98 & 89.17 & -3.06 & [-4.34, -1.85] & 7.34 \\
Modality shuffle (3 seeds) & 79.66 & 90.16 & -2.07 & [-3.09, -1.07] & 6.27 \\
\bottomrule
\end{tabular*}

\end{table}

\paragraph{Open-output contract.}
All 475 participant-facing outputs are schema-valid, with no repair, retry, or hard-gate failure; seven review warnings remain. Organizer pre-evaluation on 425 open cases reported GT 1.859/4 and VA 2.774/4. No gate intervened and the evaluator is unavailable; this establishes output-contract validity only, not clinical benefit.

\section{Discussion and Conclusion}
\label{sec:discussion}
Expert top-2 provides the principal controlled routing gain: 39 net correct cases (+2.78 percentage points) and oracle coverage of 96.15\% rather than 90.31\%. This headroom enables the final 92.23\% retrospective result; \cref{fig:rescue} illustrates the mechanism, not aggregate validity.

Cyclic relabeling is prior work; probability and log aggregation are equivalent here, and agreement is diagnostic but neutral for routing. Evidence remains retrospective, with answer regularity, image/template overlap, record-level OOF, and no patient/study IDs. Raw-image omission retains 88.95\%, so image dependence cannot establish faithful grounding. Hidden results and clinician review are unavailable.

Candidate expansion, not extra uncertainty features, supplies the main gain; open guards enforce serialization only. Future work should freeze the system for blind patient-disjoint evaluation, use set-valued targets, and compare guarded and raw answers with clinician-supported metrics.

\clearpage
\noindent\textbf{Acknowledgments.} This research was funded by the National Foundation for Science and Technology Development (NAFOSTED) through Project No. IZVSZ2\_229539 (2025--2027).

\noindent\textbf{Disclosure of Interests.} The authors have no competing interests to declare.

\bibliographystyle{splncs04}
\bibliography{references}
\end{document}